\documentclass[runningheads]{llncs}
\usepackage[T1]{fontenc}
\usepackage{bbm}
\usepackage{bm}
\usepackage{graphicx}
\usepackage{amssymb}
\usepackage{graphicx}
\usepackage{multirow}
\usepackage{booktabs}
\usepackage{amsmath}
\usepackage{color}
\usepackage{bbding}
\usepackage[colorlinks,
linkcolor=red,       
anchorcolor=blue,  
citecolor=blue,        
]{hyperref}

\begin{document}
\title{Self-aware and Cross-sample Prototypical Learning for Semi-supervised Medical Image Segmentation}
%
%
\author{Zhenxi Zhang\inst{1} \and
	Ran Ran\inst{2,3} \and
	Chunna Tian\inst{1} \and Heng Zhou \inst{1} \and Xin Li\inst{4} \and Fan Yang\inst{4} \and Zhicheng Jiao\inst{5}}
%
%
\institute{Xidian University, 2 South Taibai Road, Shanxi Province \email{zxzhang\_5@stu.xidian.edu.cn} \and
	Cancer Center, The First Affiliated Hospital of Xi'an Jiaotong University \and Precision Medicine Center, The First Affiliated Hospital of Xi'an Jiaotong University \and AIQ, Abu Dhabi, United Arab Emirates
	\and
	Department of Diagnostic Imaging, Warren Alpert Medical School of Brown University\\
}
%

\authorrunning{Zhenxi Zhang et al.}
%
%
\maketitle              
\begin{abstract}
Consistency learning plays a crucial role in semi-supervised medical image segmentation as it enables the effective utilization of limited annotated data while leveraging the abundance of unannotated data. The effectiveness and efficiency of consistency learning are challenged by prediction diversity and training stability, which are often overlooked by existing studies. Meanwhile, the limited quantity of labeled data for training often proves inadequate for formulating intra-class compactness and inter-class discrepancy of pseudo labels. 
To address these issues, we propose a self-aware and cross-sample prototypical learning method (SCP-Net) to enhance the diversity of prediction in consistency learning by utilizing a broader range of semantic information derived from multiple inputs. Furthermore, we introduce a self-aware consistency learning method which exploits unlabeled data to improve the compactness of pseudo labels within each class. Moreover, a dual loss re-weighting method is integrated into the cross-sample prototypical consistency learning method to improve the reliability and stability of our model. Extensive experiments on ACDC dataset and PROMISE12 dataset validate that SCP-Net outperforms other state-of-the-art semi-supervised segmentation methods and achieves significant performance gains compared to the limited supervised training. Our code will come soon.  
\keywords{Prototypical learning  \and Consistency learning \and Semi-supervised segmentation.}
\end{abstract}
\section{Introduction}
With the increasing demand for accurate and efficient medical image analysis, Semi-supervised segmentation methods offer a viable solution to tackle the problems associated with scarce labeled data and mitigate the reliance on manual expert annotation. It is often not feasible to annotate all images in a dataset. By exploring the information contained in the unlabeled data, semi-supervised learning \cite{zhu2009introduction,sedai2019uncertainty} can help to improve segmentation performance compared to using only a small set of annotated examples. 

Consistency constraint is a widely-used solution in semi-supervised segmentation to improve performance by making the prediction and/or intermediate features remain consistent under different perturbations. 
However, it’s challenging to obtain universal and appropriate perturbations (e.g., augmentation \cite{li2020transformation}, contexts \cite{lai2021semi}, and decoders \cite{wu2021semi}) across different tasks.
In addition, the efficacy of the consistency loss utilized in semi-supervised segmentation models could be weakened by minor perturbations that have no discernible effect on the predicted results.
Conversely, unsuitable perturbations or unclear boundaries between structures could introduce inaccurate supervisory signals, causing a build-up of errors and leading to sub-optimal performance of the model. 
Recently, some unsupervised prototypical segmentation methods \cite{xu2022all,wu2022exploring,zhang2021prototypical} apply the feature matching operation to generate the pseudo labels in the semi-supervised segmentation task. Then, the consistency constraint is enforced between the model's prediction and the corresponding prototypical prediction to enhance the model's performance. For example, Xu, et al. \cite{xu2022all} propose a cyclic prototype consistency learning framework which involves a two-way flow of information between labeled and unlabeled data. Wu, et al. \cite{wu2022exploring} suggest to facilitate the convergence of class-specific features towards their corresponding high-quality prototypes by promoting their alignment. Zhang, et al. \cite{zhang2021prototypical} exploit the feature distances from prototypes to facilitate online correction of the pseudo label in the training course.
Limited by the quantity of prototypes, the global category prototypes used in \cite{xu2022all,wu2022exploring,zhang2021prototypical} might omit diversity and impair the representation capability.

To put it briefly, prior research has neglected to consider the robustness and variability of prediction results in response to perturbations. To address this, unlike the global prototypes in \cite{xu2022all,wu2022exploring}, we introduce self-aware and cross-sample class prototypes, which generate two distinct prototype predictions to enhance semantic information interaction and ensure disagreement in consistency training. We also use prediction uncertainty between self-aware prototype prediction and multiple predictions to re-weight the consistency constraint loss of cross-sample prototypes. By doing so, we can reduce the adverse effects of label noise in challenging areas such as low-contrast regions or adhesive edges, resulting in a more stable consistency constraint training process. This, in turn, would lead to significantly improved model performance and accuracy.
Lastly, we present SCP-Net, a parameter-free semi-supervised segmentation framework that incorporates both types of prototypical consistency constraints.  

The main contributions of this paper can be summarized as follows:
(1) We conduct an in-depth study on prototype-based semi-supervised segmentation methods and propose self-aware prototype prediction and cross-sample prototype prediction to ensure appropriate prediction diversity in consistency learning. 
(2) To enhance the intra-class compactness of pseudo labels, we propose a self-aware prototypical consistency learning method. 
(3) To boost the stability and reliability of cross-sample prototypical consistency learning, we design a dual loss re-weighting method which helps to reduce the negative effect of noisy pseudo labels. 
(4) Extensive experiments on ACDC and PROMISE12 datasets have demonstrated that SCP-Net effectively utilizes the unlabeled data and improves semi-supervised segmentation performance with an annotation ratio as low as 10\%.

\section{Method}
In the semi-supervised segmentation task, the training set is divided into the labeled set $\mathcal{D}_l=\left\{\left(x_k, y_k\right)\right\}_{k=1}^{N_l}$ 
and the unlabeled set $\mathcal{D}_u=\left\{x_k\right\}_{k=N_l+1}^{N_l+N_u}$, where $N_u \gg N_l$. 
Each labeled image $ x_k \in \mathbbm{R}^{H \times W}$ has its ground-truth mask $ y_k \in \left\{0,1\right\}^{C \times H \times W} $, where 
$ H $, $ W $, and $ C $ are the height, width, and class number, respectively. Our objective is to enhance the segmentation performance of the model by extracting additional knowledge from the unlabeled dataset $\mathcal{D}_u$.

\subsection{Self-Cross Prototypical Prediction}
The prototype in segmentation refers to the aggregated representation that captures the common characteristics of some pixel-wise features from a particular object or class. 
Let $p_{ki}^{c}$ denote the probability of pixel $i$ belonging to class $c$, $f_k \in \mathbbm{R}^{D\times H \times W}$ represent the feature map of sample $k$. 
The class-wise prototypes $\bm{q_{k }^{c}}$ is defined as follows:
\begin{equation}
\label{prototype}
\bm{q_{k }^{c}}=\frac{\sum_{i} p_{k}^c(i)\cdot f_{k }(i)}{\sum_{i}p_{k}^{c}(i) }
\end{equation}
Let $B$ denote the batch size.
In the iterative training process, one mini-batch contains $B\times C$ prototypes for sample $k=1$ and other samples with index $j=2,3,\cdots, B$. Then, feature similarity is calculated according to the self-aware prototype $\bm{q_{k}^c}$ or cross-sample prototypes $\bm{q_{j}^c}$ to form multiple segmentation probability matrices. 
Specifically, $\hat{s}_{kk}^{c}$ is the self-aware prototypical similarity map via calculating the cosine similarity between the feature map $f_k$ and the prototype vector $\boldsymbol{q_{k}^{c}}$ as Eq. \ref{cossim}: 
\begin{equation}
\label{cossim}
\hat{s}_{kk}^{c}=\frac{f_{k} \cdot \boldsymbol{q_{k}^{c}}}{\left\|f_{k}\right\| \cdot\left\|\boldsymbol{q_{k}^{c}}\right\|}
\end{equation}
Then, $softmax$ function is applied to  generate the self-aware probability prediction $\hat{p}_{kk}\in \mathbbm{R}^{C \times H \times W}$ based on $\hat{s}_{kk} \in \mathbbm{R}^{C \times H \times W} $. Since $\bm{q_{k}^{c}}$ is aggregated in sample $k$ itself, which can align $f_k$ with more homologous features,
ensuring the intra-class consistency of prediction.
Similarly, we can obtain $B-1$ cross-sample prototypical similarity maps $\hat{s}_{kj}^{c}$ following Eq. \ref{cossim1}: 
\begin{equation}
\label{cossim1}
\hat{s}_{kj}^{c}=\frac{f_{k} \cdot \boldsymbol{q_{j}^{c}}}{\left\|f_{k}\right\| \cdot\left\|\boldsymbol{q_{j}^{c}}\right\|}
\end{equation}
This step ensures that features are associated and that information is exchanged in a cross-image manner.
To enhance the reliability
of prediction, we take the multiple similarity estimations $\hat{s}_{kj}\in \mathbbm{R}^{C \times H \times W}$ into consideration and integrate them to get the cross-sample probability prediction $\hat{p}_{ko}\in \mathbbm{R}^{C \times H \times W}$:
\begin{equation}
\label{multiplepro}
\hat{p}_{ko}^c=\frac{ \sum_{j=2}^{B} e^{\hat{s}_{kj}^{c}}}{\sum_{c}  \sum_{j=2}^{B} e^{\hat{s}_{kj}^{c}}}
\end{equation}
\begin{figure}[h]
	\centering
	\includegraphics[width=0.87\textwidth]{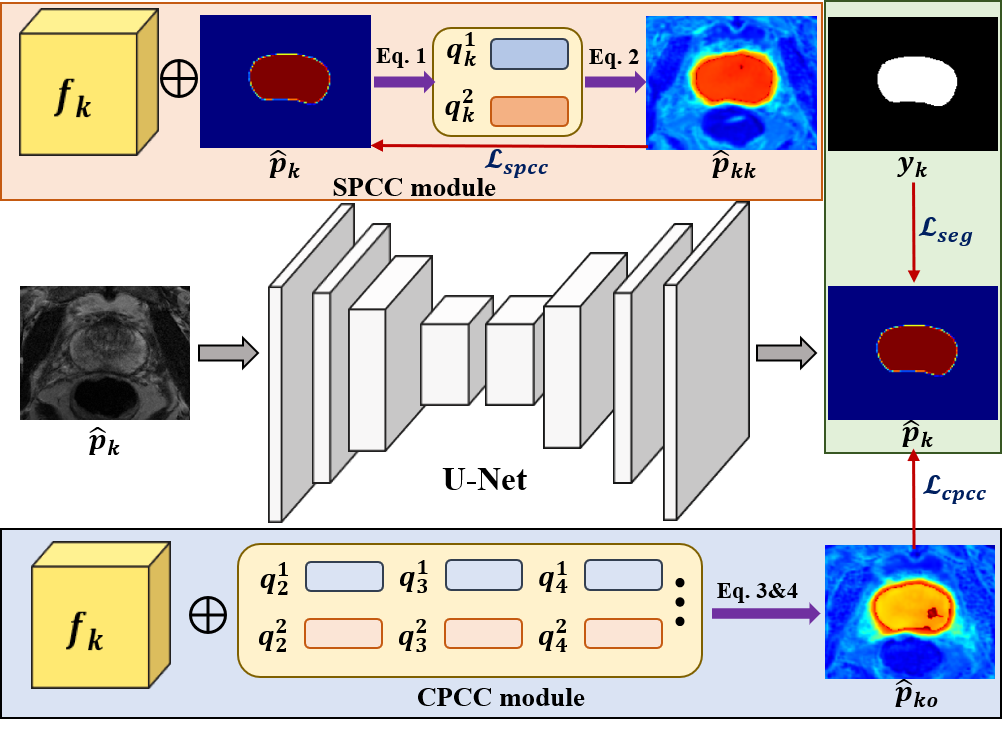}
	\caption{The overall flowchart of SCP-Net, which consists of three parts: the supervised training with $\mathcal{L}_{seg}$, SPCC module with $\mathcal{L}_{spcc}$, CPCC module with $\mathcal{L}_{cpcc}$. } \label{fig1}
\end{figure}
\subsection{Protypical Prediction Uncertainty}
To effectively evaluate the predication consistency and training stability in semi-supervised settings, we propose a prototypical prediction uncertainty estimation method based on the similarity matrices $\hat{s}_{kk}$ and $\hat{s}_{kj}$. First, we generate $B$ binary represented mask $\hat{m}_{kn} \in \mathbbm{R}^{C \times H \times W} $  via $argmax$ operation and one-hot encoding operation, where $n=1,2,\cdots,B$. Then, we sum all masks $\hat{m}_{kn}$ and dividing it by $B$ to get a normalized probability $\hat{p}_{norm}$ as:
\begin{equation}
\label{multiplepro}
\hat{p}_{norm}^c=\frac{ \sum_{n=1}^{B} {\hat{m}_{kn}^{c}}}{B}
\end{equation}
And a normalized entropy is estimated from $\hat{p}_{norm}$, denoted as ${e}_{k} \in \mathbbm{R}^{H \times W}$:
\begin{equation}
{e}_{k}=-\frac{1}{\log (C)}\sum_{c=1}^{C} \hat{p}_{norm}^{c} \log \hat{p}_{norm}^{c}
\end{equation}
where ${e}_{k}$ serves as the overall confidence of multiple prototypical predictions, and a higher entropy equals more prediction uncertainty. Then, we use  ${e}_{k} \in $ to adjust the pixel-wise weight of labeled and unlabeled samples, which will be elaborated in next subsection.
\subsection{Unsupervised Prototypical Consistency Constraint}
To enhance the prediction diversity and training effectiveness in consistency learning and mitigate the negative effect of noisy predictions in $\hat{p}_{kk}$ and $\hat{p}_{kj}$, we propose two unsupervised prototypical consistency constraints (PCC) in SPC-Net benefiting from the self-aware prototypical prediction $\hat{p}_{kk}$, cross-sample prototypical prediction $\hat{p}_{kj}$, and the corresponding uncertainty estimation $e_k$. 

\noindent\textbf{Self-aware Prototypical Consistency Constraint (SPCC)} To boost the intra-class compactness of segmentation prediction, we propose a SPCC method which applies $\hat{p}_{kk}$ as pseudo-label supervision. Therefore, the loss function of SPCC is formulated as:
\begin{equation}
\label{SPCC}
\mathcal{L}_{spcc}= \frac{1}{C\times H \times W} \sum_{i=1}^{H \times W} \sum_{c=1}^C \left\|\hat{p}_{kk}^c(i)-p_{k}^c(i)\right\|_{2}
\end{equation}
\noindent\textbf{Cross-aware Prototypical Consistency Constraint (CPCC)}
To derive dependable knowledge from other training samples, we propose a dual-weighting method for CPCC. First, we take the uncertainty estimation $e_k$ into account, which reflects the prediction stability. A higher value of $e_k$ indicates that pseudo labels with greater uncertainty may be more susceptible to errors. However, these regions provide valuable information for segmentation performance.
To reduce the influence of the suspicious pseudo labels and adjust the contribution of these crucial supervisory signals during training, we incorporate $e_k$ in CPCC by setting a weight $w_{1ki} = 1-e_{ki}$.
Second, we introduce the self-aware probability prediction $\hat{p}_{kk}$ into the CPCC module. 
Specifically, we calculate the maximum value of $\hat{p}_{kk}$ along class $c$, termed as the self-aware confidence weight $w_{2ki}$:
\begin{equation}
w_{2ki} = \max_{c}{\hat{p}_{kk}^{c}(i)}
\end{equation}
$w_{2k}$ can further enhance the reliability of CPCC. Therefore, the optimized function of CPCC is calculated between cross-aware prototypical prediction $\hat{p}_{ko}$ and $\hat{p}_k$:
\begin{equation}
\mathcal{L}_{cpcc}=\frac{1}{C\times H \times W} \sum_{i=1}^{H \times W} \sum_{c=1}^C w_{1ki} \cdot w_{2ki} \cdot \left\|\hat{p}_{ko}^c(i)-p_{k}^c(i)\right\|_{2}
\end{equation}
\noindent\textbf{Loss Function of SCP-Net}
We use the combination of cross-entropy loss $\mathcal{L}_{ce}$ and Dice loss $\mathcal{L}_{Dice}$ to supervise the training process of labeled set \cite{milletari2016v}, which is defined as:
\begin{equation}
\mathcal{L}_{seg} = \mathcal{L}_{ce}(\hat{p}_k,y_k) + \mathcal{L}_{Dice} (\hat{p}_k,y_k)
\end{equation}
For both labeled data and unlabeled data, we leverage $\mathcal{L}_{spcc}$ and $\mathcal{L}_{cpcc}$ to provide unsupervised consistency constraints for network training and explore the valuable unlabeled knowledge.
To sum it up, the overall loss function of SCPNet is the combination of the supervised loss and the unsupervised consistency loss, which is formulated as:
\begin{equation}
\mathcal{L}_{total} = \sum_{k=1}^{N_l}\mathcal{L}_{seg}(\hat{p}_k,y_k) + \lambda \sum_{k=1}^{N_l+N_u}\left(\mathcal{L}_{spcc}(\hat{p}_k,\hat{p}_{kk}) + \mathcal{L}_{cpcc}(\hat{p}_k,\hat{p}_{ko})\right)
\end{equation} 
$\lambda \left(t\right)=0.1 \cdot e^{-5\left(1-t / t_{max }\right)^2}$ is a weight using a time-dependent Gaussian warming up function \cite{tarvainen2017mean} to balance the supervised loss and unsupervised loss. 
$t$ represents the current training iteration, and $t_{max}$ is the total iterations.
\section{Experiments and Results}
\noindent\textbf{Dataset and Evaluation Metric.}
We validate the effectiveness of our method on two public benchmarks, namely the Automated Cardiac Diagnosis Challenge \footnote[1]{\url{https://www.creatis.insa-lyon.fr/Challenge/acdc/databases.html}} (ACDC) dataset \cite{bernard2018deep} and the Prostate MR Image Segmentation  challenge \footnote[2]{\url{https://promise12.grand-challenge.org}} (PROMISE12) dataset \cite{litjens2014evaluation}.
ACDC dataset contains 200 annotated short-axis cardiac cine-MRI scans from 100 subjects. All scans are randomly divided into 140 training scans, 20 validation scans, and 40 test scans following the previous work \cite{liu2022semi}.
PROMISE12 dataset contains 50 T2-weighted MRI scans which are divided into 35 training cases, 5 validation cases, and 10 test cases.
All 3D scans are converted into 2D slices. Then, each slice is resized to $256\times 256$ and normalized to $[0,1]$.
To evaluate the semi-supervised segmentation performance, we use two commonly-used evaluation metrics, the Dice Similarity Coefficient (DSC) and the Average Symmetric Surface Distance (ASSD).

\noindent\textbf{Implementation Details.}
Our method adopts U-Net \cite{ronneberger2015u} as the baseline. We use the stochastic gradient descent (SGD) optimizer with an initial learning rate of 0.1, and apply the “poly” learning rate policy to update the learning rate during training. The batch size is set to 24. Each batch includes 12 labeled slices and 12 unlabeled slices. To alleviate overfitting, we employ random flipping and random rotation to augment data. All comparison experiments and ablation experiments follow the same setup for a fair comparison, we use the same experimental setup for all comparison and ablation experiments.
All frameworks are implemented with PyTorch and conducted on a computer with a 3.0 GHz CPU, 128 GB RAM, and four NVIDIA GeForce RTX 3090 GPUs.
\begin{table}[h]
	\caption{Comparision with other methods on the ACDC test set. DSC (\%) and ASSD (mm) are reported with 14 labeled scans and 126 unlabeled scans for semi-supervised training. The bold font represents the best performance.}
	\resizebox{\linewidth}{!}{
	\begin{tabular}{@{}c|cc|cc|cc|cc|cc@{}}
		\hline
		\multirow{2}{*}{Method} & \multicolumn{2}{c|}{Scans Used} & \multicolumn{2}{c|}{RV}                                           & \multicolumn{2}{c|}{Myo}                                         & \multicolumn{2}{c|}{LV}                                          & \multicolumn{2}{c}{Avg} \\ 
		& Labeled       & Unlabeled       & DSC $\uparrow$     & ASSD $\downarrow$                            & DSC$\uparrow$                             & ASSD$\downarrow$                           & DSC$\uparrow$                             & ASSD$\downarrow$                           & DSC$\uparrow$         & ASSD$\downarrow$      \\ \hline
		U-Net                   & 14 (10\%)           & 0               & 82.24                           & 2.18                            & 80.98                           & 2.21                           & 86.89                           & 1.75                           & 83.37       & 1.60      \\
		U-Net                   & 140 (100\%)           & 0               & 91.48                           & 0.47                            & 89.22                           & 0.54                           & 94.64                           & 0.55                           & 91.78       & 0.52      \\ \hline
		MT \cite{tarvainen2017mean}                      & 14            & 126             & 87.47                           & 0.42                            & 86.19                           & 1.11                           & 90.23                           & 2.56                           & 87.97       & 1.37      \\
		UAMT \cite{yu2019uncertainty}               & 14            & 126             & 87.69                           & 0.43                            & 85.97                           & 0.76                           & 90.67                           & 2.18                           & 88.11       & 1.52      \\
		CCT \cite{ouali2020semi}                     & 14            & 126             & 87.97                           & 0.45                            & 86.07                           & 1.30                           & 89.60                           & 3.38                           & 87.88       & 1.71      \\
		URPC \cite{luo2021efficient}                  & 14            & 126             & 80.55                           & \textbf{0.39} & 84.09                           & 1.82                           & 88.76                           & 3.74                           & 84.47       & 1.98      \\
		SSNet \cite{wu2022exploring}                  & 14            & 126             & 87.21                           & 0.45                            & 86.00                           & 1.68                           & 90.91                           & 1.67                           & 88.04       & 0.97      \\
		MC-Net \cite{wu2021semi}                 & 14            & 126             & 82.69                           & 0.96                            & 84.15                           & 1.66                           & 88.86                           & 3.66                           & 85.24       & 2.09      \\
		SLC-Net \cite{liu2022semi}                 & 14            & 126             & 82.19                           & 1.93                            & 82.57                           & 1.21                           & 88.97                           & 1.25                           & 84.58       & 1.47      \\
		SCP-Net (Ours)                 & 14            & 126             & \textbf{89.26} & 0.77                            & \textbf{87.11} & \textbf{0.51} & \textbf{92.70} & \textbf{0.92} & \textbf{89.69}       & \textbf{0.73}      \\ \hline
	\end{tabular}} \label{table1}
\end{table}
\begin{figure}[h]
	\centering
	\includegraphics[width=1\textwidth]{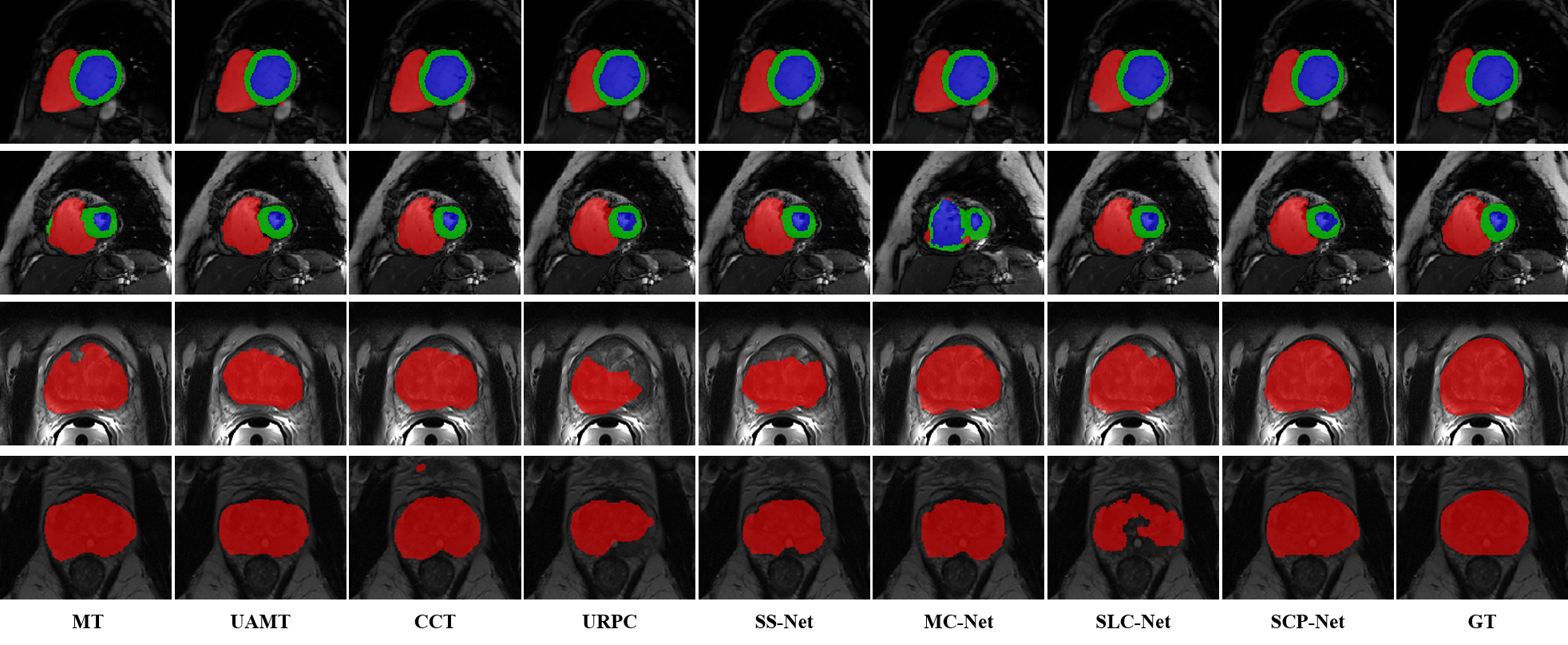}
	\caption{Visualized segmentation results of different methods on ACDC and PROMISE12. SCP-Net better preserves anatomical morphology compared to others.} 
	\label{fig2}
\end{figure}

\noindent\textbf{Comparision with Other Methods.}
To demonstrate the effectiveness of SCPNet, we compare it with 7 state-of-the-art methods for semi-supervised segmentation and fully-supervised (100\% labeled ratio) limited supervised (10\% labeled ratio) baseline. The quantitative analysis results of ACDC dataset are shown in Table \ref{table1}. SCP-Net significantly outperforms the limited supervised baseline by 7.02\%, 6.13\%, and 6.32\% on DSC for RV, Myo, and LV, respectively. SCP-Net achieves comparable DSC and ASSD to the fully supervised baseline. (89.69\% vs 91.78 and 0.73 vs 0.52). Compared with other methods, SCP-Net achieves the best DSC and ASSD, which is 1.58\% and 0.24 higher than the second-best metric, respectively. Moreover, we visualize several segmentation examples of ACDC dataset in Fig. \ref{fig2}. SCP-Net yields consistent and accurate segmentation results for the RV, Myo, and LV classes according to ground truth (GT), proving that the unsupervised prototypical consistency constraints effectively extract valuable unlabeled information for segmentation performance improvement. Table \textcolor{red}{3} in supplementary material reports the quantitative result for prostate segmentation. We also perform the limited supervised and fully supervised training with 10\% labeled ratio and 100\% labeled ratio, respectively. SCP-Net surpasses the limited supervised baseline by 16.18\% on DSC, and 10.35 on ASSD. In addition, SCP-Net gains the highest DSC of 77.06\%, which is 5.63\% higher than the second-best CCT. All improvements suggest that SPCC and CPCC are beneficial for exploiting unlabeled information.
We also visualize some prostate segmentation examples in the last two rows of Fig. \ref{fig2}. We can observe that SCP-Net generates anatomically-plausible results for prostate segmentation.

\begin{table}[h]
	\centering
	\caption{Abaliton study of the key design of SCP-Net. w means with and w/o means without. }
	\begin{tabular}{@{}c|cc|cc|cc@{}}
		\hline
		\multirow{2}{*}{Loss Function}      & \multicolumn{2}{c|}{Scans Used} & \multicolumn{2}{c|}{Weight} & \multirow{2}{*}{DSC$\uparrow$} & \multirow{2}{*}{ASSD   $\downarrow$} \\
		& Labeled       & Unlabeled       & $w_1$             & $w_2$              &                                &                                      \\ \hline
		$\mathcal{L}_{seg}$                           & 7             & 0               & w/o                & w/o                &60.88                              &13.87                                      \\
		$\mathcal{L}_{seg}$ + $\mathcal{L}_{spcc}$              & 7             & 28              & w/o                & w/o                &73.48                                &5.06                                      \\
		$\mathcal{L}_{seg}$ + $\mathcal{L}_{cpcc}$              & 7             & 28              & w                  & w                  &73.52                                &4.98                                      \\ \hline
		$\mathcal{L}_{seg}$ + $\mathcal{L}_{cpcc}$ + $\mathcal{L}_{spcc}$ & 7             & 28              & w/o                & w/o                & 74.99                                &4.05                                      \\
		$\mathcal{L}_{seg}$ + $\mathcal{L}_{cpcc}$ + $\mathcal{L}_{spcc}$ & 7             & 28              & w                  & w/o                & 76.12                              &3.78                                      \\
		$\mathcal{L}_{seg}$ + $\mathcal{L}_{cpcc}$ + $\mathcal{L}_{spcc}$ & 7             & 28              & w                  & w                  & 77.06                               &3.52                                      \\ \hline
	\end{tabular}	\label{ablstudy}
\end{table}

\begin{figure}[h]
	\centering
	\includegraphics[width=0.9\textwidth]{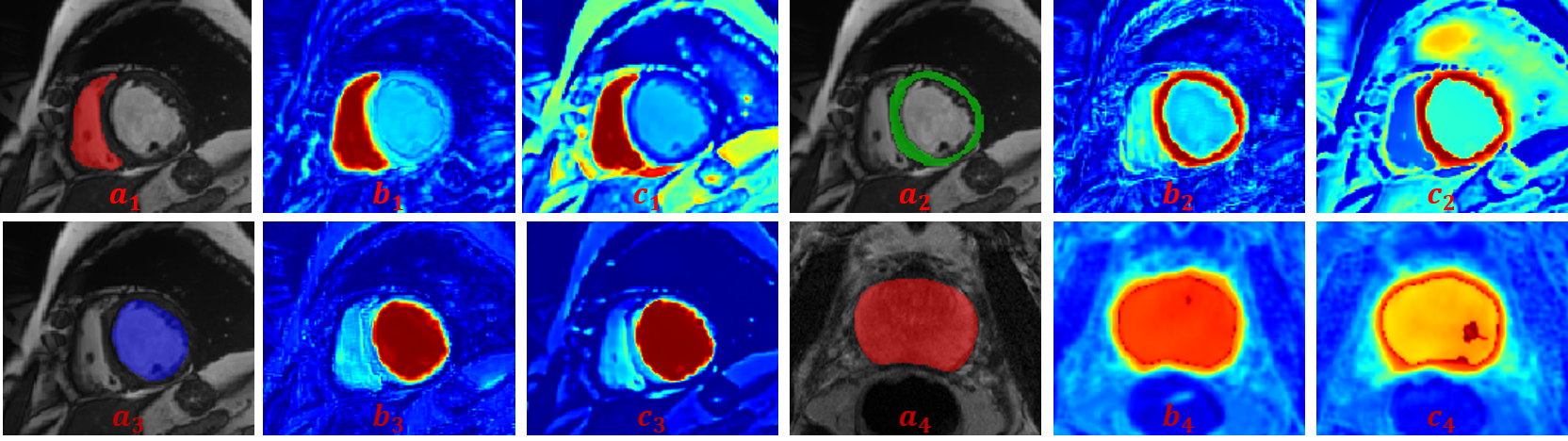}
	\caption{Visualized results for prototypical probability predictions for RV, Myo, LV, and prostate class: (a) Ground truth, (b) Self-aware probability prediction, $\hat{p}_{kk}$, (c) Cross-aware probability prediction, $\hat{p}_{ko}$. } \label{fig3}
\end{figure}
\noindent\textbf{Ablation Study.}
To demonstrate the effectiveness of the key design of SCP-Net, we perform ablation study on PROMISE12 dataset by gradually adding loss components. 
Table \ref{ablstudy} reports the results of ablation results. It can be observed that both the design of SPCC and CPCC promote the semi-supervised segmentation performance according to the first three rows, which demonstrates that PCC extracts valuable information from the image itself and other images, making them well-suited for semi-supervised segmentation. We also visualize the prototypical prediction $\hat{p}_{kk}$ and $\hat{p}_{ko}$ for different structures in Fig. \ref{fig3}. These predictions are consistent with the ground truths and show intra-class compactness and inter-class discrepancy, which validates that PCC provides effective supervision for semi-supervised segmentation. 
In the last three rows, the gradually improving performance verifies that the integration of prediction uncertainty $w_1$ and self-aware confidence $w_2$ in CPCC improves the reliability and stability of consistency training. 
\section{Conclusion}
To summarize, our proposed SCP-Net, which leverages self-aware and cross-sample prototypical consistency learning, has successfully tackled the challenges of prediction diversity and training effectiveness in semi-supervised consistency learning. The intra-class compactness of pseudo label is boosted by SPCC. The dual loss re-weighting method of CPCC enhances the model's reliability. The superior segmentation performance demonstrates that SCP-Net effectively exploits the useful unlabeled information to improve segmentation performance given limited annotated data. Moving forward, our focus will be on investigating the feasibility of learning an adaptable number of prototypes that can effectively handle varying levels of category complexity. By doing so, we expect to enhance the quality of prototypical predictions and improve the overall performance.

\clearpage
\subsubsection{Supplementary Materials}
In this section, we report the result of comparative experiments on the PROMISE12 test set in Table \ref{comPro}. 

\begin{figure}[h]
	\centering
	\includegraphics[width=0.9\textwidth]{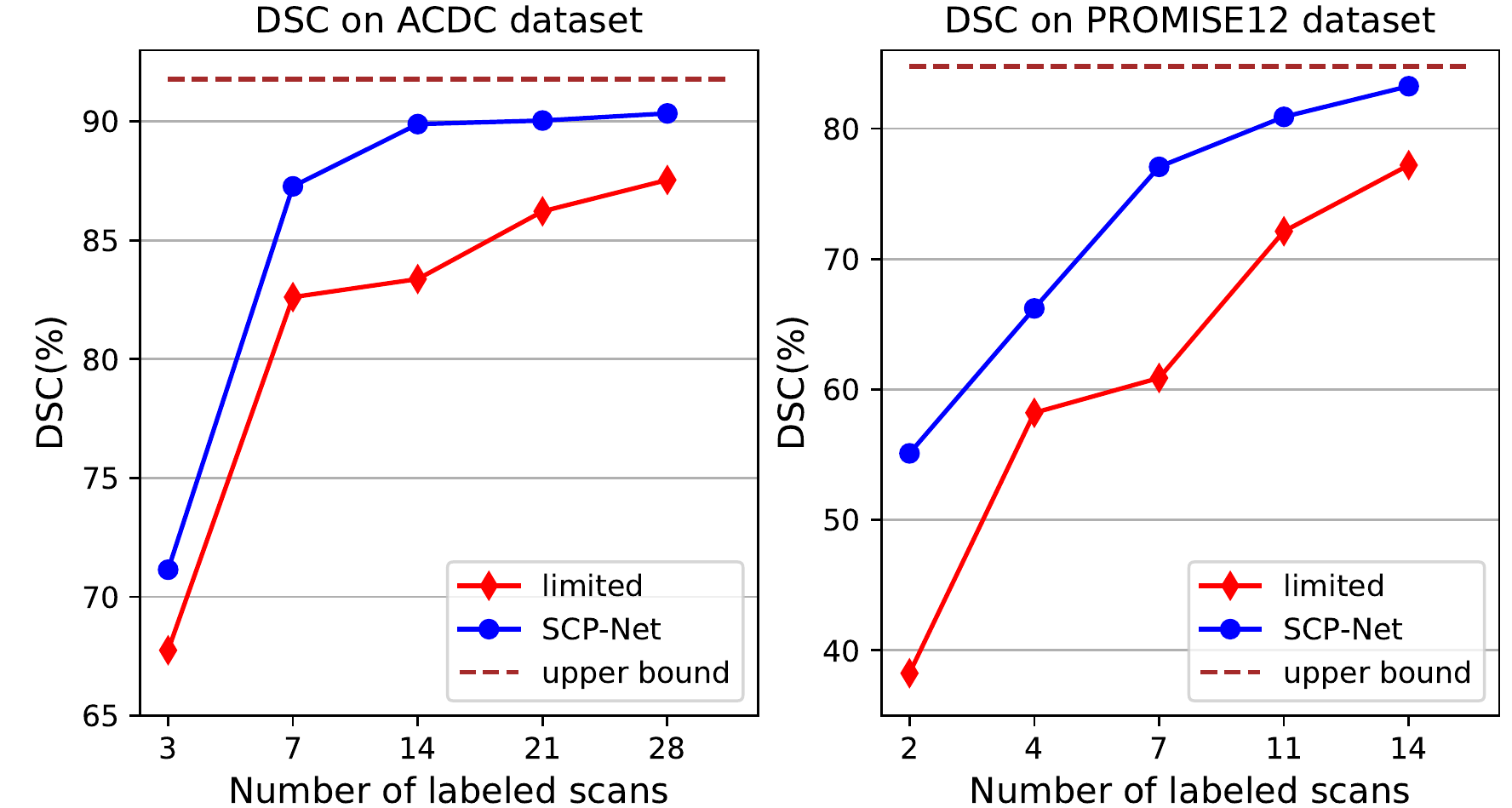}
	\caption{Segmentation performance with different data setting on ACDC and PROMISE12 dataset. ``limited'' represent the limited supervised baseline. The upper bound performance is obtained by training with 100\% labeled data.} \label{differentratio}
\end{figure}

\begin{table}[h]
	\centering
	\caption{Comparision with other methods on the PROMISE12 test set. The bold font represents the best evaluation metric.}
	
	\begin{tabular}{@{}c|cc|cc@{}}
		\hline
		\multirow{2}{*}{Method} & \multicolumn{2}{c|}{Scans Used} & \multirow{2}{*}{DSC$\uparrow$} & \multirow{2}{*}{ASSD$\downarrow$} \\ 
		& Labeled         & Unlabeled     &                      &                       \\ \hline
		U-Net                   & 7 (10\%)        & 0             & 60.88                & 13.87                 \\
		U-Net                   & 35 (100\%)      & 0             & 84.76                & 1.58                  \\ \hline
		MT \cite{tarvainen2017mean}                     & 7               & 28            & 71.42                & 7.58                  \\
		UAMT \cite{yu2019uncertainty}                    & 7               & 28            & 65.69                & \textbf{2.43}                  \\
		CCT \cite{ouali2020semi}                    & 7               & 28            & 71.43                & 16.61                 \\
		URPC \cite{luo2021efficient}                  & 7               & 28            & 63.23                & 4.33                  \\
		SSNet \cite{wu2022exploring}                  & 7               & 28            & 62.31                & 4.36                  \\
		MC-Net \cite{wu2021semi}                  & 7               & 28            & 67.36                & {2.81}                  \\
		SLC-Net \cite{liu2022semi}                 & 7               & 28            & 68.31                & 4.69                  \\
		SCP-Net (Ours)                    & 7               & 28            & \textbf{77.06}                & 3.52                  \\ \hline
	\end{tabular}\label{comPro}
\end{table}
Moreover, we report the segmentation results of SCP-Net when using a different number of labeled scans in Fig. \ref{differentratio}. 
As shown in Fig. \ref{differentratio} and Table \ref{diffpro},  SCP-Net achieves consistent significant improvements compared with the corresponding limited supervised baseline which only uses labeled scans in the training process for prostate segmentation. Notably, when the split of labeled/unlabeled is 14/21 on PROMISE12 dataset, SCP-Net gains close performance to the upper bound performance which is achieved by the fully supervised baseline (DSC: 83.24\% vs 84.76\%). Similar results can be observed in Fig. \ref{differentratio} and Table \ref{diffACDC} on ACDC dataset. Specifically, SCP-Net outperforms the corresponding limited supervised baseline consistently under different splits and achieves comparable results to the fully-supervised baseline with 28 labeled scans and 112 unlabeled scans (DSC: 90.34\% vs 91.78\%). All these results suggest that SCP-Net successfully exploits the useful information from unlabeled data by self-aware and cross-sample prototypical consistency learning. Our code will be released on GitHub after the review of this paper. 

\begin{table}[h]
	
	\centering
	\caption{Segmentation results of SCP-Net using different labeled scans compared with the limited supervised baseline on ACDC dataset.}
	\begin{tabular}{@{}l|cc|cc|cc|cc|cc@{}}
		\hline
		\multirow{2}{*}{Method} & \multicolumn{2}{c|}{Scans Used} & \multicolumn{2}{c|}{RV}                & \multicolumn{2}{c|}{Myo}                                         & \multicolumn{2}{c|}{LV}                                          & \multicolumn{2}{c}{Avg} \\
		& Labeled       & Unlabeled       & DSC$\uparrow$                             & ASSD$\downarrow$ & DSC$\uparrow$                             & ASSD$\downarrow$                           & DSC$\uparrow$                             & ASSD$\downarrow$                           & DSC$\uparrow$         & ASSD$\downarrow$      \\ \hline
		U-Net                   & 140           & 0               & \textbf{91.48}                           &\textbf{0.47} & \textbf{89.22}                           & \textbf{0.54}                           & \textbf{94.64}                           & \textbf{0.55}                           & \textbf{91.78}       & \textbf{0.52}      \\ \hline
		U-Net                   & 3             & 0               & 56.61                           & 8.50 & 69.20                           & 4.08                           & 77.46                           & 3.33                           & 67.75       & 5.30      \\
		SCP-Net                 & 3             & 137             & 60.65                           & 7.15 & 71.90                           & 2.26                           & 80.87                           & 4.55                           & 71.14       & 4.65      \\
		U-Net                   & 7             & 0               & 82.63                           & 2.06 & 78.84                           & 3.21                           & 86.37                           & 4.35                           & 82.61       & 3.21      \\
		SCP-Net                 & 7             & 133             & 86.84                           & 1.42 & 84.43                           & 2.42                           & 90.54                           & 4.10                           & 87.27       & 2.65      \\
		U-Net                   & 14            & 0               & 82.24                           & 2.18 & 80.98                           & 2.21                           & 86.89                           & 1.75                           & 83.37       & 1.60      \\
		SCP-Net                 & 14            & 126             & 89.26                           & 0.77 & 87.11                           & 0.51                           & 92.70                           & 0.92                      & 89.69       & 0.73      \\
		U-Net                   & 21            & 0               & 85.73                           & 2.90 & 83.76                           & 1.51                           & 89.18                           & 2.84                           & 86.22       & 2.41      \\
		SCP-Net                 & 21            & 119             & 89.38                           & 0.40 & 88.01                           & 0.39                           & 92.74                           & 0.62                           & 90.04       & 0.47      \\
		U-Net                   & 28            & 0            & 87.45                           & 0.69 & 85.14                           & 0.89                           & 90.03                           & 1.71                           & 87.54       & 1.09      \\
		SCP-Net                 & 28            & 112             & \textbf{89.48} & \textbf{0.34} & \textbf{88.33} & \textbf{0.45} & \textbf{93.19} & \textbf{0.61} & \textbf{90.34}       & \textbf{0.47}      \\ \bottomrule
	\end{tabular} \label{diffACDC}
\end{table}
\begin{table}[h]
	\centering
	\caption{Segmentation results of SCP-Net using different labeled scans compared with the limited supervised baseline on PROMISE12 dataset.}
	\begin{tabular}{@{}lcccc@{}}
		\toprule
		\multirow{2}{*}{Method} & \multicolumn{2}{c}{Scans Used} & \multirow{2}{*}{DSC$\uparrow$} & \multirow{2}{*}{ASSD$\downarrow$} \\ 
		& Labeled       & Unlabeled      &                      &                       \\ \midrule
		U-Net                   & 35            & 0              & \textbf{84.76}                & \textbf{1.58}                  \\ \midrule
		U-Net                   & 2             & 0              & 38.25                & 33.91                 \\
		SCP-Net                 & 2             & 33             & 55.10                & 19.45                 \\
		U-Net                   & 4             & 0              & 58.21                & 17.83                 \\
		SCP-Net                 & 4             & 31             & 66.21                & 11.56                 \\
		U-Net                   & 7             & 0              & 60.88                & 13.87                 \\
		SCP-Net                 & 7             & 28             & 77.06                & 3.52                  \\
		U-Net                   & 11            & 0              & 72.12                & 3.47                  \\
		SCP-Net                 & 11            & 24             & 80.89                & 3.13                  \\
		U-Net                   & 14            & 0              & 77.19                & 3.17                  \\
		SCP-Net                 & 14            & 21             & \textbf{83.24}                & \textbf{2.68}                  \\ \bottomrule
	\end{tabular}	\label{diffpro}
\end{table}

%
\clearpage

%
%
%
\bibliographystyle{splncs04_unsort}
\bibliography{refs}

\begin{thebibliography}{10}
\providecommand{\url}[1]{\texttt{#1}}
\providecommand{\urlprefix}{URL }
\providecommand{\doi}[1]{https://doi.org/#1}

\bibitem{zhu2009introduction}
Zhu, X., Goldberg, A.B.: Introduction to semi-supervised learning. Synthesis
  Lectures on Artificial Intelligence and Machine Learning  \textbf{3}(1),
  1--130 (2009)

\bibitem{sedai2019uncertainty}
Sedai, S., Antony, B., Rai, R., Jones, K., Ishikawa, H., Schuman, J., Gadi, W.,
  Garnavi, R.: Uncertainty guided semi-supervised segmentation of retinal
  layers in oct images. In: Medical Image Computing and Computer Assisted
  Intervention--MICCAI 2019: 22nd International Conference, Shenzhen, China,
  October 13--17, 2019, Proceedings, Part I 22. pp. 282--290. Springer (2019)

\bibitem{li2020transformation}
Li, X., Yu, L., Chen, H., Fu, C.W., Xing, L., Heng, P.A.:
  Transformation-consistent self-ensembling model for semisupervised medical
  image segmentation. IEEE Transactions on Neural Networks and Learning Systems
   \textbf{32}(2),  523--534 (2020)

\bibitem{lai2021semi}
Lai, X., Tian, Z., Jiang, L., Liu, S., Zhao, H., Wang, L., Jia, J.:
  Semi-supervised semantic segmentation with directional context-aware
  consistency. In: Proceedings of the IEEE/CVF Conference on Computer Vision
  and Pattern Recognition. pp. 1205--1214 (2021)

\bibitem{wu2021semi}
Wu, Y., Xu, M., Ge, Z., Cai, J., Zhang, L.: Semi-supervised left atrium
  segmentation with mutual consistency training. In: International Conference
  on Medical Image Computing and Computer-Assisted Intervention. pp. 297--306.
  Springer (2021)

\bibitem{xu2022all}
Xu, Z., Wang, Y., Lu, D., Yu, L., Yan, J., Luo, J., Ma, K., Zheng, Y., Tong,
  R.K.y.: All-around real label supervision: Cyclic prototype consistency
  learning for semi-supervised medical image segmentation. IEEE Journal of
  Biomedical and Health Informatics  \textbf{26}(7),  3174--3184 (2022)

\bibitem{wu2022exploring}
Wu, Y., Wu, Z., Wu, Q., Ge, Z., Cai, J.: Exploring smoothness and
  class-separation for semi-supervised medical image segmentation. In: Medical
  Image Computing and Computer Assisted Intervention--MICCAI 2022: 25th
  International Conference, Singapore, September 18--22, 2022, Proceedings,
  Part V. pp. 34--43. Springer (2022)

\bibitem{zhang2021prototypical}
Zhang, P., Zhang, B., Zhang, T., Chen, D., Wang, Y., Wen, F.: Prototypical
  pseudo label denoising and target structure learning for domain adaptive
  semantic segmentation. In: Proceedings of the IEEE/CVF conference on computer
  vision and pattern recognition. pp. 12414--12424 (2021)

\bibitem{milletari2016v}
Milletari, F., Navab, N., Ahmadi, S.A.: V-net: Fully convolutional neural
  networks for volumetric medical image segmentation. In: 2016 Fourth
  International Conference on 3D Vision (3DV). pp. 565--571. IEEE (2016)

\bibitem{tarvainen2017mean}
Tarvainen, A., Valpola, H.: Mean teachers are better role models:
  Weight-averaged consistency targets improve semi-supervised deep learning
  results. Advances in Neural Information Processing Systems  \textbf{30}
  (2017)

\bibitem{bernard2018deep}
Bernard, O., Lalande, A., Zotti, C., Cervenansky, F., Yang, X., Heng, P.A.,
  Cetin, I., Lekadir, K., Camara, O., Ballester, M.A.G., et~al.: Deep learning
  techniques for automatic mri cardiac multi-structures segmentation and
  diagnosis: is the problem solved? IEEE Transactions on Medical Imaging
  \textbf{37}(11),  2514--2525 (2018)

\bibitem{litjens2014evaluation}
Litjens, G., Toth, R., Van De~Ven, W., Hoeks, C., Kerkstra, S., van Ginneken,
  B., Vincent, G., Guillard, G., Birbeck, N., Zhang, J., et~al.: Evaluation of
  prostate segmentation algorithms for mri: the promise12 challenge. Medical
  Image Analysis  \textbf{18}(2),  359--373 (2014)

\bibitem{liu2022semi}
Liu, J., Desrosiers, C., Zhou, Y.: Semi-supervised medical image segmentation
  using cross-model pseudo-supervision with shape awareness and local context
  constraints. In: Medical Image Computing and Computer Assisted
  Intervention--MICCAI 2022: 25th International Conference, Singapore,
  September 18--22, 2022, Proceedings, Part VIII. pp. 140--150. Springer (2022)

\bibitem{ronneberger2015u}
Ronneberger, O., Fischer, P., Brox, T.: U-net: Convolutional networks for
  biomedical image segmentation. In: Medical Image Computing and
  Computer-Assisted Intervention--MICCAI 2015: 18th International Conference,
  Munich, Germany, October 5-9, 2015, Proceedings, Part III 18. pp. 234--241.
  Springer (2015)

\bibitem{yu2019uncertainty}
Yu, L., Wang, S., Li, X., Fu, C.W., Heng, P.A.: Uncertainty-aware
  self-ensembling model for semi-supervised 3d left atrium segmentation. In:
  International Conference on Medical Image Computing and Computer-Assisted
  Intervention. pp. 605--613. Springer (2019)

\bibitem{ouali2020semi}
Ouali, Y., Hudelot, C., Tami, M.: Semi-supervised semantic segmentation with
  cross-consistency training. In: Proceedings of the IEEE/CVF Conference on
  Computer Vision and Pattern Recognition. pp. 12674--12684 (2020)

\bibitem{luo2021efficient}
Luo, X., Liao, W., Chen, J., Song, T., Chen, Y., Zhang, S., Chen, N., Wang, G.,
  Zhang, S.: Efficient semi-supervised gross target volume of nasopharyngeal
  carcinoma segmentation via uncertainty rectified pyramid consistency. In:
  Medical Image Computing and Computer Assisted Intervention--MICCAI 2021: 24th
  International Conference, Strasbourg, France, September 27--October 1, 2021,
  Proceedings, Part II 24. pp. 318--329. Springer (2021)

\end{thebibliography}
\end{document}